\documentclass[]{fairmeta}

\usepackage{pgfplots}
\pgfplotsset{compat=1.18}
\usepackage{sansmath}
\usepackage{nicematrix}
\usepackage{colortbl}
\usepackage{xcolor}
\usepackage{tabularx}
\usepackage{wrapfig}
\usepackage{booktabs}
\usepackage{nicematrix}
\usepackage{graphicx}
\usepackage{subcaption}
\usepackage{amssymb}
\usepackage{adjustbox}
\usepackage{xspace}
\usepackage{tabularx}
\usepackage{makecell}
\newcommand{\eg}{\emph{e.g.}\xspace}
\newcommand{\ie}{\emph{i.e.}\xspace}
\newcommand{\best}[1]{\cellcolor{green!20}\textbf{#1}}
\newcommand{\modelname}{SAM 3D Body}
\newcommand{\modelcode}{3DB}
\newcommand{\meshname}{Momentum Human Rig}
\newcommand{\meshcode}{MHR}

\title{SAM 3D Body: Robust Full-Body Human Mesh Recovery}

\author[\star]{Xitong Yang}
\author[\star]{Devansh Kukreja}
\author[\star]{Don Pinkus}

\author[]{Anushka Sagar}
\author[]{Taosha Fan}
\author[\circ]{Jinhyung Park}
\author[\circ]{Soyong Shin}
\author[]{Jinkun Cao}
\author[]{Jiawei Liu}
\author[]{Nicolas Ugrinovic}

\author[\dagger]{Matt Feiszli}
\author[\dagger]{Jitendra Malik}
\author[\dagger]{Piotr Dollar}
\author[\dagger]{Kris Kitani}

\affiliation{Meta Superintelligence Labs}

\contribution[\star]{Core Contributor}
\contribution[\circ]{Intern}
\contribution[\dagger]{Project Lead}

\abstract{
    We introduce \modelname{} (\modelcode), a promptable model for single-image full-body 3D human mesh recovery (HMR) that demonstrates state-of-the-art performance, with strong generalization and consistent accuracy in diverse in-the-wild conditions.
    \modelcode{} estimates the human pose of the body, feet, and hands. It is the first model to use a new parametric mesh representation, \meshname{} (\meshcode{}), which decouples skeletal structure and surface shape. \modelcode{} employs an encoder–decoder architecture and supports auxiliary prompts, including 2D keypoints and masks, enabling user-guided inference similar to the SAM family of models. We derive high-quality annotations from a multi-stage annotation pipeline that uses various combinations of manual keypoint annotation, differentiable optimization, multi-view geometry, and dense keypoint detection. Our data engine efficiently selects and processes data to ensure data diversity, collecting unusual poses and rare imaging conditions. We present a new evaluation dataset organized by pose and appearance categories, enabling nuanced analysis of model behavior. Our experiments demonstrate superior generalization and substantial improvements over prior methods in both qualitative user preference studies and traditional quantitative analysis. Both \modelcode{} and \meshcode{} are open-source.
}

\metadata[Demo]{\url{https://www.aidemos.meta.com/segment-anything/editor/convert-body-to-3d}}
\metadata[Code]{\url{https://github.com/facebookresearch/sam-3d-body}}
\metadata[Website]{\url{https://ai.meta.com/sam3d}}

\begin{document}

\maketitle

  \begin{center}
      \centering
      \includegraphics[width=0.95\textwidth]{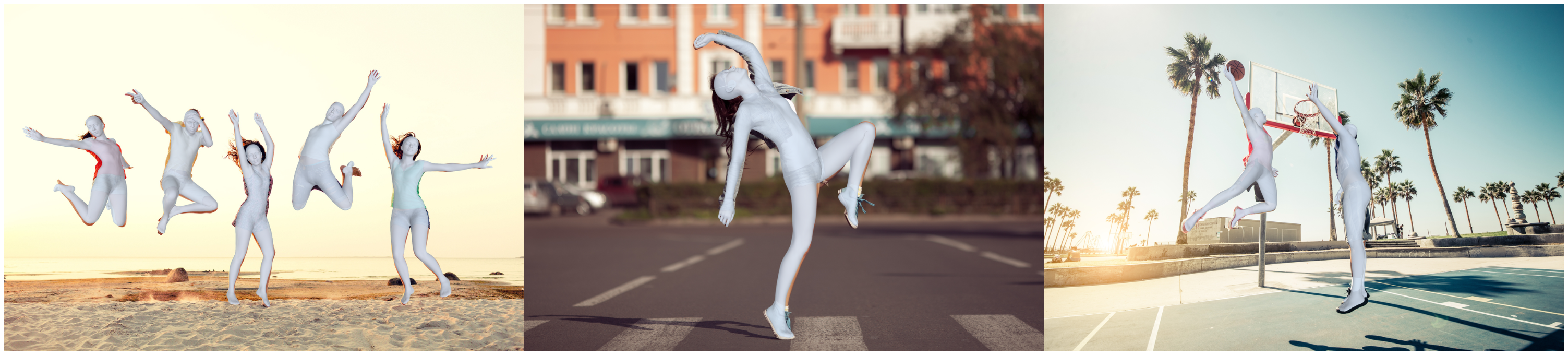}%
      \captionof{figure}{Full-body human mesh recovery results using \modelname{} (\modelcode{}). Our model demonstrates robust performance in estimating challenging poses across diverse viewpoints and produces
  accurate body and hand pose estimations within a unified framework.}
      \label{fig:wide}
  \end{center}

\section{Introduction}
\label{section:intro}

Estimating 3D human pose (skeleton pose and structure) and shape (soft body tissue) from images is an essential capability for vision and embodied AI systems to understand and interact with people. Despite notable progress in human mesh recovery (HMR)~\cite{goel2023hmr20, dwivedi2024tokenhmr, patel2024camerahmr, wang2025blade, wang2025prompthmr}, existing approaches still exhibit unsatisfactory robustness when applied to in-the-wild images, which limits their applicability to real-world scenarios such as robotics~\cite{peng2018sfv, patel2022learning, vasilopoulos2020reactive} and biomechanics~\cite{pearl2023fusion}. In particular, current models often fail on individuals presenting challenging poses, severe occlusion, or captured from uncommon viewpoints. They also struggle to reliably estimate both the overall body pose and the fine details of the hands and feet in a unified full-body framework. 

We argue that the primary challenges in developing a robust full-body human mesh recovery model stem from both the data and model aspects. First, collecting large-scale and diverse human pose datasets with high-quality mesh annotations is inherently difficult and computationally costly. Most existing datasets either suffer from low diversity due to laboratory capture settings~\cite{chao2021dexycb,h36m_pami,Joo_2017_TPAMI} or from low mesh quality resulting from pseudo-labeling~\cite{vonmarcard20183dpw,andriluka2014mpii}. Second, current HMR architectures do not adequately address the distinct optimization mechanisms required for body and hand pose estimation, nor do they incorporate effective training strategies to handle uncertainty and ambiguity from monocular images.

In this work, we present \modelname{} (\modelcode{}), a robust full-body HMR model fueled by large-scale, high-quality human pose data curated by our data engine. 

\noindent \textbf{Robust Full-body HMR Model.} We make three main contributions to improve model performance on both body and hand pose estimation. (i) We propose a novel promptable encoder–decoder architecture~\cite{kirillov2023sa1b, ravi2025sam2} that enables the model to condition on optional 2D keypoints, masks or camera information for controllable pose estimation. This promptable design naturally facilitates interactive guidance in ambiguous or challenging scenarios during training, and provides a coherent approach to integrate hand and body predictions.
(ii) Our model utilizes a shared image encoder and two separate decoders for the body and hands. This two-way-decoder design effectively alleviates conflicts in optimizing body and hand pose estimation, which arise from differences in input resolution, camera estimation, and supervision objectives.
(iii) Unlike most prior work that relies on the SMPL~\cite{loper2015smpl} human mesh model, we build \modelcode{} on a new parametric mesh representation, \meshcode{}~\cite{stoll2025momentumhumanrig}, which decouples skeletal pose and body shape, providing richer control and interpretability for full-body reconstruction.

\noindent \textbf{Data Engine for Diverse Human Pose and High-quality Annotation.} HMR methods have increasingly turned to large-scale training data for higher performance~\cite{goel2023hmr20, cai2023smplerx, yin2025smplestx}. However, high-quality 3D supervision remains scarce, and existing in-the-wild datasets are still limited in scale and diversity. To this end, we design a new data creation pipeline that features: 
(i) \textit{Data Quality}: Our annotation pipeline combines various combinations of components such as geometric constraints, parametric priors, and dense keypoint regression, which automatically yields high-quality 3D human mesh annotations.
(ii) \textit{Data Quantity}: We curate data from large licensed stock photo repositories, multiple multi-view capture datasets, and synthetic data. We create a large scale of \textbf{7 million} images with high-quality annotation. 
(iii) \textit{Data Diversity}: Our data is diversified using a VLM-based data engine that mines for in-the-wild challenging images and routes them for annotation. This ensures coverage of rare poses, difficult viewpoints, and varied appearances, providing a more diverse dataset for supervision.

Together, the data engine and full-body HMR model enable \modelcode{} to recover high-fidelity full-body human meshes from a single image. \modelcode{} achieves state-of-the-art performance across both body and hand pose estimation. Extensive experiments demonstrate that \modelcode{} consistently outperforms prior HMR methods on standard metrics, generalizes better to unseen datasets, and is preferred by users in a study of $7,800$ participants, achieving a significant $5:1$ win rate in visual quality.
To our knowledge, it is the first single model that delivers the \textbf{best performance to body-specialized models and comparable performance to hand-specialized models}, while providing interactive control and strong robustness under challenging poses and in-the-wild scenarios. 

\section{Related Work}
\label{section:related}

\noindent\textbf{Human Mesh Models:} The most widely used human mesh model is SMPL~\cite{loper2015smpl}, which parameterizes human body into pose and shape. SMPL-X~\cite{pavlakos2019smplx} goes further to include hands (MANO~\cite{romero2017mano}) and faces (FLAME~\cite{li2017flame}). SMPL models intertwine the skeletal structure and soft-tissue mass within the \emph{shape space}, which can limit interpretability (\eg, the parameters do not always map directly to bone lengths) and controllability. Alternatively, \meshname{}~\cite{stoll2025momentumhumanrig}, an enhancement of ATLAS \cite{park2025atlas}, explicitly decouples the skeletal structure and body shape, and we adopt it as our representation of the human body. 

\noindent\textbf{Human Mesh Recovery (HMR):} Early HMR methods like {HMR~2.0}~\cite{goel2023hmr20} were \emph{body-only} methods that predicted the body without articulated hands or feet ~\cite{kolotouros2019learning,li2022cliff,dwivedi2024tokenhmr}. Instead, \modelcode{} follows the more recent paradigm of full-body methods~\cite{baradel2024multihmr,choutas2020expose,rong2021frankmocap,cai2023smplerx,wang2025prompthmr} that estimate \emph{body+hands+feet}. There are also part-specific hand mesh recovery methods~\cite{pavlakos2023hamer,potamias2025wilor} that only estimate the pose and shape of the hands, which usually have more accurate performance compared to full-body methods. In contrast, \modelcode{} shows strong performance on both hand and full-body estimation.

\noindent\textbf{Promptable Inference:} Promptable inference, popularized by the SAM family~\cite{kirillov2023sa1b, ravi2025sam2}, enables user or system-provided prompts (such as 2D keypoints or masks) to guide model predictions. Similarly to \cite{wang2025prompthmr}, our approach supports various prompt types, including 2D keypoints and masks, and by integrating prompt tokens directly into the transformer architecture, enables user-guided mesh recovery.

\noindent\textbf{Data Quality and Annotation Pipelines:} A major bottleneck in HMR is the quality of training data. Many datasets rely on pseudo-ground-truth (pGT) meshes obtained from monocular fitting~\cite{kolotouros2019learning, kanazawa2018end}, which often contain systematic errors in pose, shape, and camera parameters~\cite{patel2024camerahmr}. Recent work~\cite{dwivedi2024tokenhmr, wang2025blade} highlights the impact of annotation noise on reported metrics and generalization. To address this, multi-view datasets~\cite{goliath2025, khirodkar2024harmony4d, moon2020interhand} and synthetic data have been used in our work to provide higher-fidelity supervision. Our method builds on these insights by employing a scalable data engine that mines challenging cases using vision-language models, and by leveraging a multi-stage annotation pipeline that combines dense keypoint detection, strong parametric priors, and robust optimization.

\begin{figure}[tb]
  \centering
  \includegraphics[width=1.0\columnwidth]{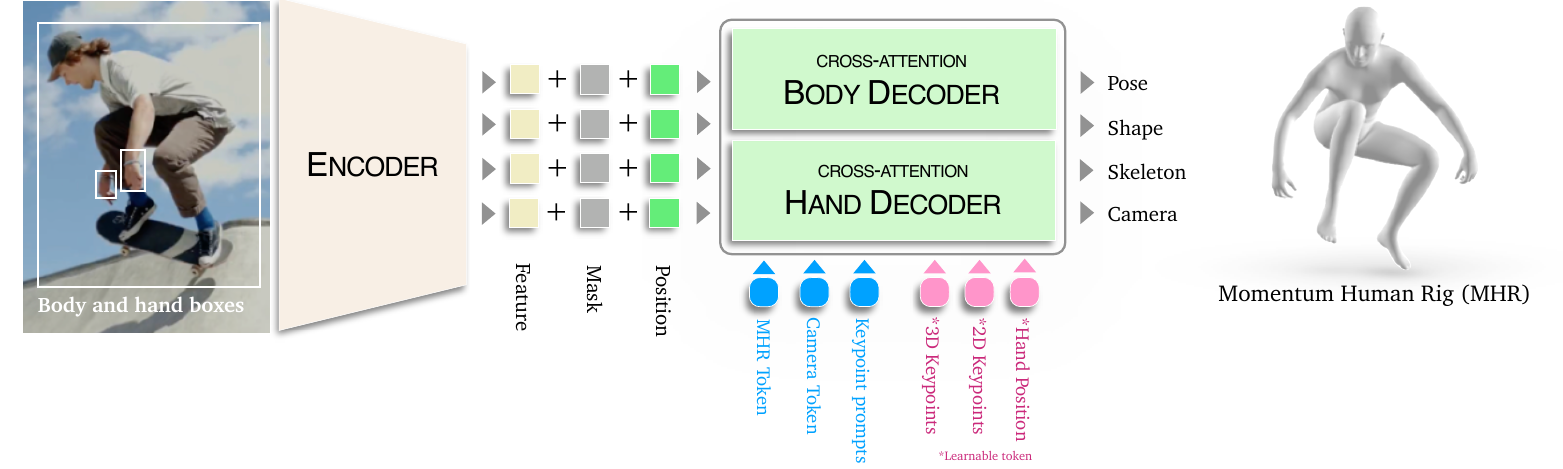}
  \caption{\modelname{} Model Architecture. We employ a promptable encoder–decoder architecture with a shared image encoder and separate decoders for body and hand pose estimation.}
  \label{fig:model}
\end{figure}

\section{\modelname{} Model Architecture}
Our goal is to recover 3D human meshes (\ie, \meshcode{} parameters) accurately, robustly and interactively from a single image. To this end, we design \modelcode{} as a promptable encoder–decoder architecture (see \Cref{fig:model}) with a rich set of prompt tokens. \modelcode{} is designed to be \emph{interactive} as it can accept 2D keypoints or masks, allowing users or downstream systems to guide inference.

\subsection{Image Encoder}
The human-cropped image $I$ is normalized and passed through a vision backbone to produce a dense feature map $F$. An optional set of hand crops $I_{\mathrm{hand}}$ can also be provided to obtain hand crop feature maps $F_{\mathrm{hand}}$: 
\begin{align}
    F &= \mathrm{ImgEncoder}(I), \\
    F_{\mathrm{hand}} &= \mathrm{ImgEncoder}(I_{\mathrm{hand}}).
\end{align}

\modelcode{} considers two optional prompts: 2D keypoints and segmentation masks. Keypoint prompts are encoded by positional encodings summed with learned embeddings and are provided as additional tokens for the pose decoder. Mask prompts are embedded using convolutions and summed element-wise with the image embedding~\cite{kirillov2023sa1b}.

\subsection{Decoder Tokens}
\modelcode{} has two decoders: The body decoder outputs the full-body human rig and an optional hand decoder can provide enhanced hand pose results. The pose decoders take a set of \emph{query tokens} as input to predict the parameters of \meshcode{} and camera parameters. There are four types of query tokens: \meshcode{}+camera, 2D keypoint prompt, auxiliary 2D/3D keypoint tokens and optional hand position tokens.

\noindent\textbf{\meshcode{}+Camera Token:} The initial estimate of \meshcode{} and (optionally) camera parameters is embedded as a learnable token for MHR parameter estimation:
\begin{align}
    T_{\mathrm{pose}} &= \mathrm{RigEncoder}(E_{\mathrm{init}}) \in \mathbb{R}^{1 \times D}, \\
    E_{\mathrm{init}} &\in \mathbb{R}^{d_{\mathrm{init}}}.
\end{align}
    
\noindent\textbf{2D Keypoint Prompt Tokens:} If 2D keypoint prompts $K$ are provided (\eg, from a user or detector), they are encoded as:
\begin{align}
    T_{\mathrm{prompt}} &= \mathrm{PromptEncoder}(K) \in \mathbb{R}^{N \times D}, \\
    K &\in \mathbb{R}^{N \times 3},
\end{align}
where each keypoint is represented by $(x, y, \text{label})$.

\noindent\textbf{Hand Position Tokens:} The hand token, $T_{\mathrm{hand}} \in \mathbb{R}^{2 \times D}$, is used in the body decoder to locate the hand positions inside the human images.
This set of tokens is optional, without which \modelcode{} can still produce a full-body human rig because the output from body decoder already includes hands.

\noindent\textbf{Auxiliary Keypoint Tokens:} To further enhance interactivity and model capacity, we include learnable tokens for all 2D and 3D keypoints.
\begin{align}
    T_{\mathrm{keypoint2D}} &\in \mathbb{R}^{J_{2D} \times D}, \\
    T_{\mathrm{keypoint3D}} &\in \mathbb{R}^{J_{3D} \times D}.
\end{align}
These tokens allow the model to reason about specific joints and support downstream tasks such as keypoint prediction or uncertainty estimation.

\subsection{MHR Decoder} 
All tokens are concatenated to form the full set of queries:
\begin{equation}
    T = \left[ T_{\mathrm{pose}},\; T_{\mathrm{prompt}},\; T_{\mathrm{keypoint2D}},\; T_{\mathrm{keypoint3D}},\; T_{\mathrm{hand}}\right]
\end{equation}
This flexible assembly enables the model to operate in both fully automatic and user-guided modes, adapting to the available prompts. The body decoder attends to both the query tokens $T$, the full-body image features $F$, 
%and the left and right hand image features $F_L$ and $F_R$:
\begin{equation}
O  = \mathrm{Decoder}(T,F) \in \mathbb{R}^{(3 + N + J_{2D} + J_{3D}) \times D}.
\end{equation}
Through cross-attention, the body decoder fuses prompt information with visual context, enabling robust and editable mesh recovery. Optionally, the hand decoder can take the same prompt information while attends to the hand crop features $F_{\mathrm{hand}}$ to provide another output token $O_{\mathrm{hand}}$.

% \noindent\textbf{\meshcode{} Output Format:} 
% For \meshcode{}, the shape parameters $\mathbf{S_s} \in \mathbb{R}^{45}$ are coefficients over decoupled head, body, and hand vertex linear bases. The skeletal parameters $\mathbf{S_k} \in \mathbb{R}^{28}$ determine bone lengths of major body and hand joints. \meshcode{} is articulated by full-body pose parameters $\mathbf{P} \in \mathbb{R}^{130}$. \modelcode{} also predicts the camera rotation and translation $\mathbf{C} \in \mathbb{R}^{6}$ for precise image alignment of the mesh.
The first output token of $O$ is passed through an MLP to regress the final mesh parameters: $\theta = \mathrm{MLP}(O_0) \in \mathbb{R}^{d_{\mathrm{out}}}$, where $\theta = \{ \mathbf{P}, \mathbf{S}, \mathbf{C}, \mathbf{S}_k \}$ are the predicted \meshcode{} parameters: pose, shape, camera pose and skeleton, respectively. Another set of outputs can be computed from $O_{\mathrm{hand}}$ for a pair of MHR hands, which can be merged to the body output to improve the estimation of the hand.

\section{Model Training and Inference} 

\noindent\textbf{Model Training.} \modelcode{} is trained with a comprehensive multi-task loss terms, $\mathcal{L}_{\text{train}} = \sum_{i} \lambda_i \mathcal{L}_i$, where each $\mathcal{L}_i$ is a task-specific loss targeting a specific prediction head or anatomical structure. $\lambda_i$ are hyper-parameters set empirically. To stabilize training, certain loss terms (\eg, 3D keypoints) are introduced with a warm-up schedule, gradually increasing their weights over the course of training. We also simulate an interactive setup~\cite{kirillov2023sa1b,sofiiuk2022reviving} for training by randomly sampling prompts in multiple rounds per sample. This multi-task, prompt-aware loss design provides strong supervision across all outputs. We describe the losses in details below.

\noindent\textbf{2D/3D Keypoint Loss:} We supervise 2D/3D joint locations using an $L_1$ loss, incorporating learnable per-joint uncertainty to modulate the loss based on prediction confidence. For 3D body and hand keypoints, we normalize them with their respective pelvis and wrist locations before computing the loss. Hand keypoints are weighted according to annotation availability. 2D keypoints are supervised in the cropped image spaces, and we upweight the loss for the user-provided keypoint to encourage prompt consistency when keypoint prompts are available. 

\noindent\textbf{Parameter Losses:} MHR parameters (pose, shape) are supervised with $L_2$ regression losses, and joint limit penalties are imposed to discourage anatomically implausible poses. %Notably, for improved hand alignment, we separately supervise the global orientation of the wrist, as opposed to only local pose supervision.
%To ensure accurate surface geometry, we apply an $L_1$ loss between predicted and ground-truth mesh vertices, including both the full-resolution mesh and lower-resolution variants (\eg, 595-vertex subsets). 

\noindent\textbf{Hand Detection Loss:} \modelcode{} can localize the hand position by a built-in hand detector. We apply GIoU loss and $L_1$ loss to supervise the hand box regression. We also predict the uncertainty of hand boxes and turn off the hand decoder on hand-occluded samples during inference. 

\noindent\textbf{Full-body Inference.}
\label{sec:inf}
During inference, we use the body decoder output by default, with the option to merge hand decoder output when hands are detected.
The benefit of the hand decoder comes from the hand-specific data used during training and the flexibility of a free-moving wrist due to the dedicated prediction head. By unifying the hand decoder's output to the body decoder, our model can provide full-body prediction with improved hand pose estimation.
However, we noticed that simply integrating the hand decoder's output into the middle of the \meshname{} kinematic tree can lead to errors in adjacent joints, particularly at the elbows. Therefore, we leverage the promptability of 3DB  to mitigate these errors introduced by the hand decoder's output. Specifically, we use the wrist location from the hand decoder as well as the elbow location predicted by the body decoder to prompt the body decoder to generate a refined full-body pose estimation result. 
Finally, the predicted local \meshcode{} parameters are merged to a full-body configuration following the kinematic tree of the mesh model. We show qualitative comparisons of our strategy in ~\Cref{fig:ablation_qual}.

\section{Data Engine for Diversity}
\label{sec:de}
Obtaining highly accurate human mesh annotations paired with the images can be computationally costly. Instead, one common strategy is to annotate a large video collection and leveraging temporal constraints to get more reliable pseudo annotations. While it is possible to get a large number of training images from videos, the poses, appearance, imaging conditions, and background might be very similar. In order to increase the diversity of our training dataset, we implemented an automated data engine that selectively routes difficult images for annotation, enabling scalable and efficient dataset curation. 

At the core of our data engine is a Vision-Language Model (VLM) driven mining strategy. Rather than relying on simple heuristics or random sampling, we leverage VLMs to automatically generate and update mining rules that identify high-value images for annotation. The VLM identifies images exhibiting challenging scenarios for pose estimation, including occlusion (where the human subject is partially hidden by objects or other people), unusual poses (rare or complex body configurations such as acrobatics or dance), interaction (human-object or human-human activities like holding tools or group actions), extreme scale (subjects appearing at atypical distances from the camera), low visibility (poor lighting, motion blur, or partial visibility), and hand-body coordination (tight coupling of hand and body poses, as in sign language or sports).

Mining rules are automatically updated iteratively based on failure analysis of the current model, allowing the engine to adaptively focus on the most challenging or informative samples. Failure analysis is performed semi-manually, by evaluating \modelcode{} on the current set of annotated images, visualizing the most challenging images using keypoint location error, and then manually annotating the image with a few words. These words and images are used to create text prompt for the VLM. New images selected by the VLM are then routed for manual annotation. By focusing annotation efforts on the most informative samples, our data engine enables efficient search through tens of millions of images, while maximizing the value and diversity of each annotated image. By collecting a highly diverse dataset, it provides the basis on which to build a very robust HMR model that works on a wide range of in-the-wild images.

\section{Data Annotation and Mesh Fitting}
\label{sec:anno}

In addition to the robustness enabled by the data diversity derived from our data engine, the accuracy of our model depends heavily on the quality of our annotations. To this end, we designed a multi-stage annotation pipeline that produces accurate 3D mesh pseudo-ground truth from both in-the-wild single image datasets and a variety of multi-view datasets, using various combinations of manual 2D keypoint annotation, sparse and dense keypoint detection, geometric constraints, temporal constraints, strong parametric priors, and robust optimization methods.

\subsection{Manual Annotation} 
\begin{figure}[tb]
    \centering
    \includegraphics[width=1.00\linewidth]{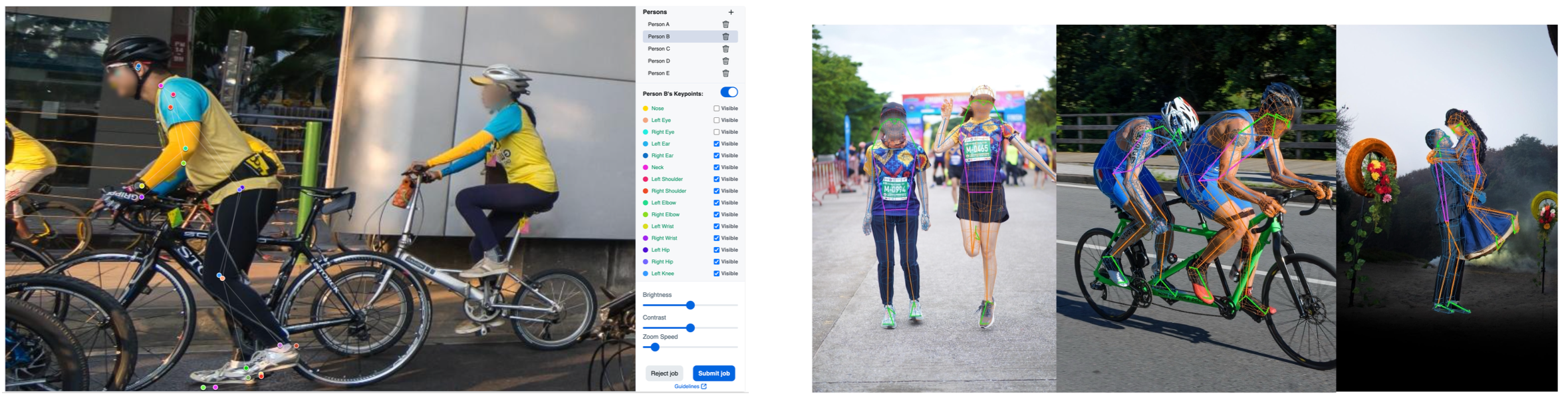}
    \caption{Left: GUI of our annotation tool for annotating 2D keypoints. Right: Comparison of the dense (thin) and sparse (thick) keypoints for pseudo annotation.}
    \label{fig:combined-figures}
\end{figure}

Given a set of images selected by the data engine, we use a current version of \modelcode{} to estimate initial 2D joint positions. Then, a team of trained annotators review and manually correct the estimated joint locations if needed, as shown in \Cref{fig:combined-figures}(a). The annotators also assign a per-joint visibility label according to a strict rubric. Joints with substantial occlusion or other factors that would prevent accurate placement (\eg, \>50\% occlusion, motion blur) are marked as \textit{not visible}.

\begin{figure}[tb]
    \centering
    \includegraphics[width=1.00\linewidth]{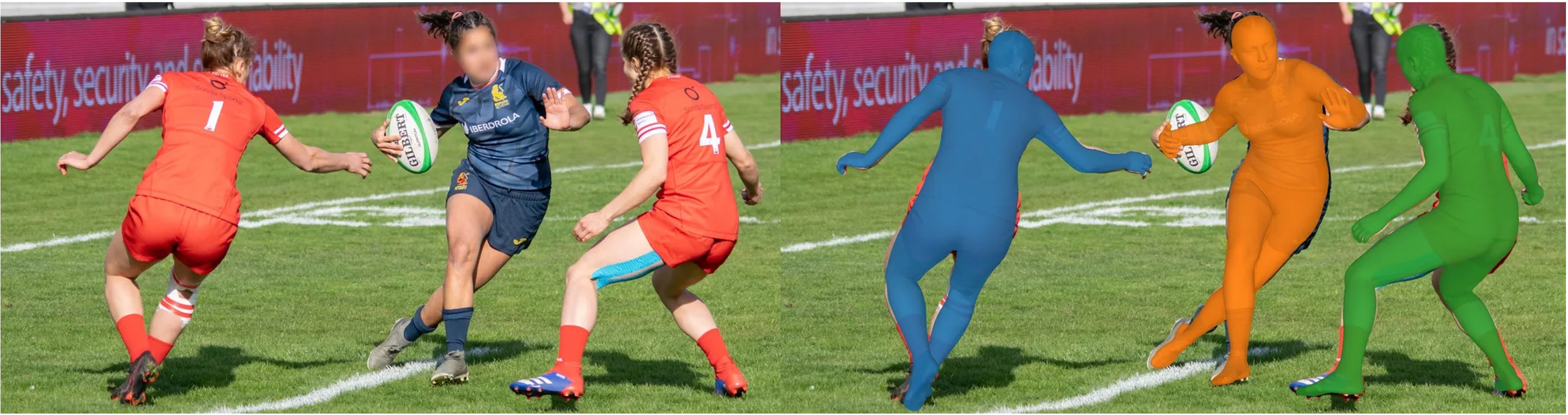}
    \caption{Example of single-image \meshcode{} mesh fitting for ITW datasets. Source: SA-1B~\cite{kirillov2023sa1b}.}
    \label{fig:img2mesh}
\end{figure}

\begin{figure}[tb]
    \centering
    \begin{subfigure}[b]{0.56\linewidth}
        \centering
        \includegraphics[width=\linewidth]{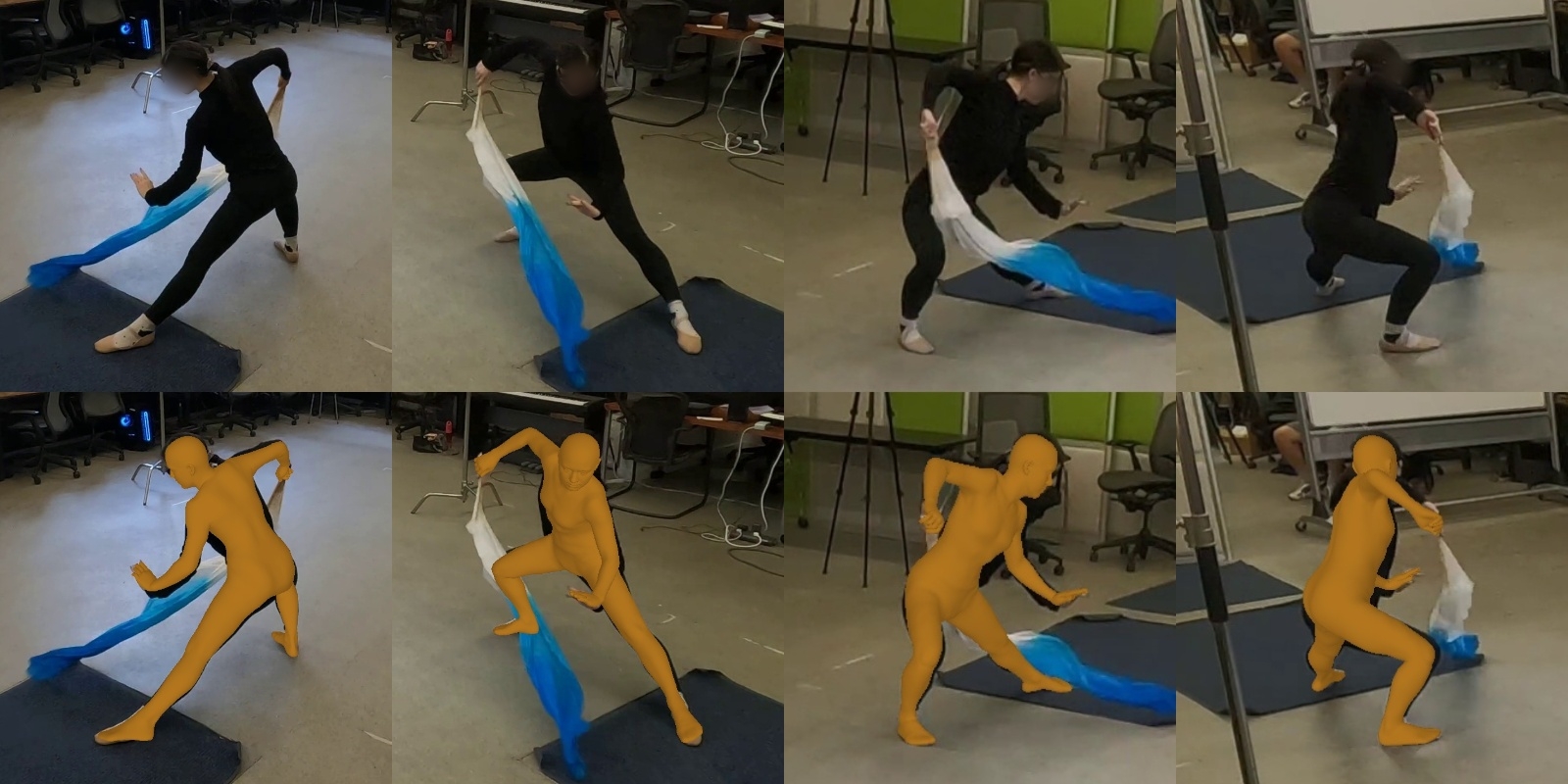}
        % \caption{Multi-view MHR mesh fitting. Source: EgoExo4D}
        \caption{}
        \label{fig:egoexo4d}
    \end{subfigure}
    \hfill
    \begin{subfigure}[b]{0.38\linewidth}
        \centering
        \includegraphics[width=\linewidth]{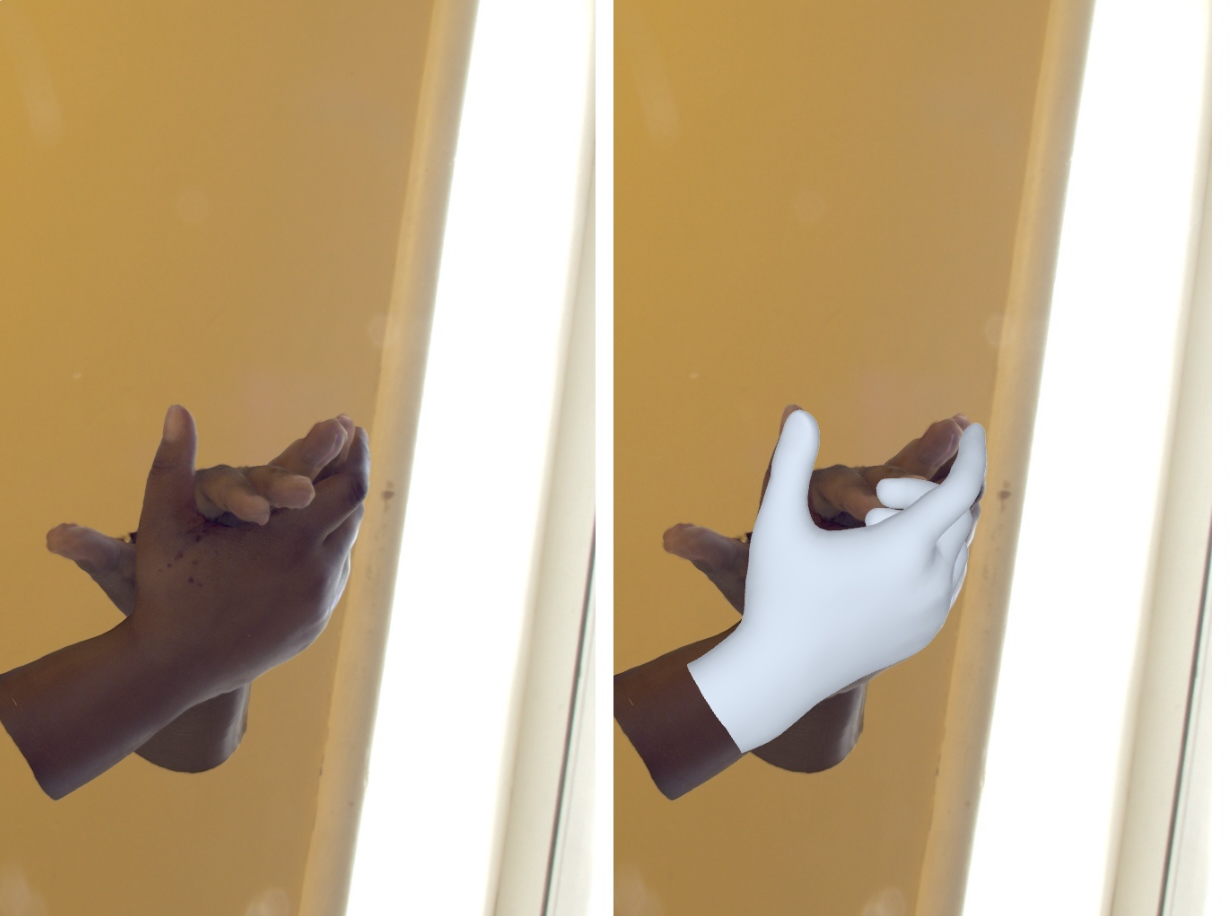}
        \caption{}
        \label{fig:dense_keypoints}
    \end{subfigure}
    \caption{Examples of MHR mesh fitting results. (a) Multi-view mesh fitting. Source: EgoExo4D~\cite{grauman2024egoexo4d}. (b) Scan-based mesh fitting. Source: Re:Interhand~\cite{mueller2023reinterhand}.}
    \label{fig:combined-figures-multiview}
\end{figure}

\subsection{Single-Image Mesh Fitting} 
For each image, we first obtain the initial estimation of \meshcode{} parameters from a current version of \modelcode's predictions, as well as the $595$ dense 2D keypoints predicted from a high-capacity keypoint detector. \meshcode{} fitting is then performed via gradient-based refinement of the model parameters, minimizing a composite fitting loss $\mathcal{L}_{\text{fit}} = \sum_{j} \lambda_j \mathcal{L}_j$, where each $\mathcal{L}_j$ is a task-specific loss including 2D keypoint loss, initialization-anchored regularization and priors. Hyper-parameters $\lambda_j$ are set via cross-validation.  We apply several loss terms and priors to make the fitting goal: \textbf{2D Keypoint Loss} is the L2 distance between projected and detected dense 2D keypoints, to ensure minimal 2D reprojection error.  \textbf{Initialization-Anchored Regularization} penalizes deviation from the initial prediction by applying L2 losses on both the \meshname{} parameters and their corresponding 3D keypoints, thereby preventing model drift. \textbf{Pose and Shape Prior} enforces anatomical plausibility via a learned Gaussian Mixture prior and L2 regularization. Following the pipeline above, we derive the image to \meshcode{} fittings as training supervision as in~\Cref{fig:img2mesh}.

\noindent\textbf{Dense keypoint detector.} The configuration of 595 dense keypoints is chosen as it represents the minimal manifold of a human body mesh for capturing diverse body shapes and hand poses. The dense keypoint detector adopts a standard Transformer encoder-decoder architecture. However, unlike prior models that only exploit visual cues from pixels \cite{patel2024camerahmr, hewitt2024look, mamma2025}, our dense keypoint detector leverages additional sparse keypoint guidance obtained from the manual annotation step to predict accurate 2D dense keypoints from in-the-wild images, as illustrated in \Cref{fig:combined-figures}(b). We first train the model on 3D datasets (\eg, Goliath and Synthetic), and use it for multi-stage mesh fitting on the in-the-wild datasets (\eg, COCO, AI Challenger, MPII). We then project the MHR mesh to dense keypoints for a second round of dense keypoint detector training and apply this iterative training scheme twice.

\subsection{Multi-View Mesh Fitting} 
Though single-view mesh fitting is effective for a large and diverse set of images, the annotation quality tends to be lower fidelity due to the depth ambiguities and natural occlusion. Therefore, we also exploit multi-view mesh fitting on suitable datasets.
For multi-view video datasets, we further extend the pipeline to jointly fit mesh across all frames and camera views, leveraging both spatial and temporal cues. Synchronized 2D keypoints are extracted for each camera and frame, then triangulated to obtain sparse 3D keypoints. 

The mesh model is initialized from these triangulated points and camera parameters and refined via second-order optimization-based update of the model parameters, minimizing a composite fitting loss, $
\mathcal{L}_{\text{multi}} = \sum_{k} \lambda_k \mathcal{L}_k,$
where each \( \mathcal{L}_k \) is a task-specific loss including \textbf{the 2D keypoint loss and the regularization and priors as single-view mesh fitting}, together with additional 3D keypoint loss and temporal smoothness: \textbf{3D Keypoint Loss} is the L2 distance between mesh joints and triangulated 3D keypoints obtained from multi-view geometry, providing strong spatial supervision.  \textbf{Temporal Smoothness Loss} encourages estimated pose parameters to temporally smooth, penalizing abrupt changes in motion and promoting realistic temporal dynamics.
 \( \lambda_k \) are set via cross-validation. Optimization alternates between updating camera parameters, shape, skeleton, and pose, with robust keypoint filtering (\eg, robust losses, RANSAC, smoothing). Body specific parameters (\eg, shape, skeleton parameters) are optimized jointly across frames. The mesh fitting happens on body full-body data and hand data as shown in~\Cref{fig:combined-figures-multiview}.

\section{Training Datasets}

We train our model on a mix of single-view, multi-view, and synthetic datasets listed in \Cref{tab:dataset_training}, covering general body pose, hands, interactions, and ``in-the-wild' conditions to ensure the quality, quantity and diversity of training data.

\noindent \textbf{Single-view in-the-wild:} We utilize datasets that captures people in unconstrained environments with diverse appearance, pose, and scene conditions. For this, we use AIChallenger~\cite{wu2019aic}, MS COCO~\cite{lin2014coco}, MPII~\cite{andriluka2014mpii}, 3DPW~\cite{vonmarcard20183dpw}, and a subset of SA-1B~\cite{kirillov2023sa1b}. 

\noindent \textbf{Multi-view consistent:} To incorporate geometric consistency for more reliable annotations, we use multi-view data from Ego-Exo4D~\cite{grauman2024egoexo4d}, Harmony4D~\cite{khirodkar2024harmony4d}, EgoHumans~\cite{khirodkar2023egohumans}, InterHand2.6M~\cite{moon2020interhand}, DexYCB~\cite{chao2021dexycb} and Goliath ~\cite{goliath2025}. 

\noindent \textbf{High-fidelity synthetic:} We use a photorealistic synthetic extension of the Goliath dataset~\cite{goliath2025}. It provides millions of frames with ground-truth \meshcode{} parameters across diverse identities, clothing, and contexts. Synthetic data ensures accurate supervision for human mesh recovery, complementing real-world datasets that prioritize diversity over quality. 

\noindent \textbf{Hand datasets:} These datasets (marked with $\star$ in \Cref{tab:dataset_training}), such as Re:Interhand~\cite{mueller2023reinterhand}, are used to train both the body and hand decoder. We provide wrist-truncated hand samples to train the hand decoder.

\begin{table}[t]
  \centering
  \small
  \caption{List of \modelcode{} training datasets. $\star$ denotes the datasets providing samples to train the hand decoder.}
  \begin{adjustbox}{width=.5\linewidth}
  \begin{NiceTabular}{lcccc}
  \textbf{Dataset} & \textbf{\# Images/Frames} & \textbf{\# Subjects} & \textbf{\# Views} \\
  \midrule
  MPII human pose~\cite{andriluka2014mpii}  & 5K      & 5K+     & 1     \\
  MS COCO~\cite{lin2014coco}                & 24K     & 24K+    & 1     \\
  3DPW~\cite{vonmarcard20183dpw}            & 17K     & 7       & 1     \\
  AIChallenger~\cite{wu2019aic}             & 172K    & 172K+   & 1     \\
  SA-1B~\cite{kirillov2023sa1b}             & 1.65M   & 1.65M+  & 1     \\
  Ego-Exo4D~\cite{grauman2024egoexo4d}      & 1.08M   & 740     & 4+    \\
  DexYCB~\cite{chao2021dexycb}              & 291K    & 10      & 8     \\
  EgoHumans~\cite{khirodkar2023egohumans}   & 272K    & 50+     & 15    \\
  Harmony4D~\cite{khirodkar2024harmony4d}   & 250K    & 24      & 20    \\
  InterHand~\cite{moon2020interhand}$\star$        & 1.09M   & 27      & 66    \\
  Re:Interhand~\cite{mueller2023reinterhand}$\star$& 1.50M & 10 & 170 \\
  Goliath~\cite{goliath2025}$\star$                & 966K    & 120+    & 500+  \\
  Synthetic$\star$                                 & 1.63M   & --      & --    \\
  \end{NiceTabular}
  \end{adjustbox}
  \label{tab:dataset_training}
\end{table}

\section{Evaluation}

We follow prior HMR work and report standard pose and shape evaluation metrics: MPJPE \cite{martinez2017simple}, PA-MPJPE \cite{zhang2020inference}, PVE \cite{li2021hybrik}, and PCK \cite{zhang2020inference}. To evaluate on SMPL-based datasets, a \meshcode{} mesh is mapped to the SMPL mesh format. We present results with two variants of the model; \modelcode{}-H leverages the commonly used ViT-H (632M) backbone, and \modelcode{}-DINOv3 uses the recent DINOv3 (840M)~\cite{simeoni2025dinov3} encoder. We resize the input to $512 \times 512$ for the image encoder and use an off-the-shelf field-of-view (FOV) estimator (\eg, MoGe-2~\cite{wang2025moge}) to provide camera intrinsics for model inference.

\subsection{Evaluating Performance on Common Datasets}

We first evaluate \modelcode{} on five standard benchmark datasets in \Cref{tab:common}, comparing with a wide variety of state-of-the-art (SoTA) mesh recovery methods. \modelcode{} outperforms all other single-image methods and is even competitive with video-based approaches that additionally leverage temporal information. 

In particular, our model achieves superior results in the EMDB and RICH datasets, which are \textit{out-of-domain} (\ie, not included in the training set), indicating better generalization than previous SoTA methods. 
\modelcode{} exceeds the second best model, NLF, on all datasets in terms of 3D metrics except for RICH which dataset NLF uses in training while our model does not.
\modelcode{} is also state-of-the-art on PCK for 2D evaluation on the COCO and LSPET datasets, demonstrating strong 2D alignment.

\begin{table*}[tb]
    \caption{Comparison on five common benchmarks. The best results are highlighted in bold, while the second-best results are underlined. Results evaluated using publicly released checkpoint denoted by $^\dagger$. Models trained using RICH denoted by $^*$.}
    \label{tab:common}
    \centering
    \resizebox{0.95\textwidth}{!}{
\begin{NiceTabular}{clcccccccccccc}
    %\toprule
    & & \multicolumn{3}{c}{3DPW (14)}  & \multicolumn{3}{c}{EMDB (24)}  & \multicolumn{3}{c}{RICH (24)}  & \multicolumn{1}{c}{COCO} & \multicolumn{1}{c}{LSPET} \\
    \cmidrule(lr){3-5} \cmidrule(lr){6-8} \cmidrule(lr){9-11} \cmidrule(lr){12-12} \cmidrule(lr){13-13}
    & Models & \scriptsize{PA-MPJPE $\downarrow$} & \scriptsize{MPJPE $\downarrow$} & \scriptsize{PVE $\downarrow$} & \scriptsize{PA-MPJPE $\downarrow$} & \scriptsize{MPJPE $\downarrow$} & \scriptsize{PVE $\downarrow$} & \scriptsize{PA-MPJPE $\downarrow$} & \scriptsize{MPJPE $\downarrow$} & \scriptsize{PVE $\downarrow$} & \scriptsize{PCK@0.05 $\uparrow$} & \scriptsize{PCK@0.05 $\uparrow$} \\
    \midrule
    \multirow{5}{1em}{\rotatebox[origin=c]{90}{\textsc{image}}}
    & HMR2.0b \cite{goel2023hmr20}        & 54.3 & 81.3 & 93.1 & 79.2 & 118.5 & 140.6 & 48.1$^\dagger$ & 96.0$^\dagger$ & 110.9$^\dagger$ & 86.1 & 53.3 \\
    & CameraHMR \cite{patel2024camerahmr}   & 35.1 & 56.0 & 65.9 & 43.3 & 70.3 & 81.7 & 34.0 & 55.7 & 64.4 & 80.5$^\dagger$ & 49.1$^\dagger$ \\
    & PromptHMR \cite{wang2025prompthmr}    & 36.1 & 58.7 & 69.4 & 41.0 & 71.7 & 84.5 & 37.3 & 56.6 & 65.5 & 79.2$^\dagger$ & 55.6$^\dagger$ \\
    & SMPLerX-H \cite{cai2023smplerx}     & 46.6$^\dagger$ & 76.7$^\dagger$ & 91.8$^\dagger$ & 64.5$^\dagger$ & 92.7$^\dagger$ & 112.0$^\dagger$ & 37.4$^\dagger$ & 62.5$^\dagger$ & 69.5$^\dagger$ & -- & -- \\
    & NLF-L+fit$^\ast$ \cite{sarandi2024nlf}       & 33.6 & 54.9 & \underline{63.7} & 40.9 & 68.4 & 80.6 & \textcolor{gray}{28.7$^\dagger$} &  \textcolor{gray}{51.0$^\dagger$} &  \textcolor{gray}{58.2$^\dagger$} & 74.9$^\dagger$ & 54.9$^\dagger$ \\
    \midrule
    \multirow{3}{1em}{\rotatebox[origin=c]{90}{\textsc{video}}}
    & WHAM \cite{shin2024wham}              & 35.9 & 57.8 & 68.7 & 50.4 &  79.7 & 94.4 &   --  &   --  &   --  &   --  &   -- \\
    & TRAM \cite{wang2024tram}              & 35.6 & 59.3 & 69.6 & 45.7 &  74.4 & 86.6 &   --  &   --  &   --  &   --  &   -- \\
    & GENMO \cite{li2025genmo}              & 34.6 & \best{53.9} & 65.8 & 42.5 &  73.0 & 84.8 & 39.1  & 66.8  & 75.4  &   --  &   -- \\
    \midrule
    & 3DB-H (Ours) & \best{33.2} & \underline{54.8}  & 64.1  & \underline{38.5} & \underline{62.9} & \underline{74.3}  & \underline{31.9} & \underline{55.0} & \underline{61.7} &  \best{86.8} & \best{68.9} \\  
    & 3DB-DINOv3 (Ours) & \underline{33.8} & \underline{54.8} & \best{63.6} & \best{38.2} & \best{61.7} & \best{72.5} & \best{30.9} & \best{53.7} & \best{60.3} & \underline{86.5} & \underline{67.8} \\
    %\bottomrule
\end{NiceTabular}
    }
\end{table*}

\subsection{Evaluating Performance on New Datasets}\label{sec:new}

Throughout our experiments, we found that mesh recovery models are particularly fragile in out-of-domain settings due to camera, appearance, and pose differences. To understand how methods perform on new, unseen data distributions, we additionally evaluate on five new datasets (38.6K images) in \Cref{tab:newdata}. The five new datasets include (1) Ego-Exo4D~\cite{grauman2024egoexo4d}, (2) Harmony4D~\cite{khirodkar2024harmony4d}, (3) Goliath~\cite{goliath2025}, (4) in-house synthetic data and (5) SA1B-Hard. Ego-Exo4D captures humans in diverse, skilled activities, divided into physical (EE4D-Phys) and procedural (EE4D-Proc) domains. Harmony4D focuses on close multi-human interaction in dynamic sports settings. Goliath offers diverse motions in a precise, studio environment. The synthetic dataset consists of single-human images with diverse camera angles and parameters. SA1B-Hard is a subset of 2.6K images extracted from SA1B using our data engine. Together, these five new datasets present a challenging new testbed for mesh recovery methods.

As it is difficult to compare methods using the exact same training data and methodology due to prohibitive data usage licenses, unclear descriptions of training data, and lack of training code (CameraHMR, PromptHMR, and NLF are trained on 6, 9, and 48 datasets, respectively), we fairly test the generalization ability of \modelcode{} by using a leave-one-out training procedure. This ensures a fair comparison with prior work which have also not seen these datasets. To serve as an in-domain, upper bound comparison, we also show the performance of \modelcode{} when trained on the \textit{full dataset} (\ie, training data is also sampled from these new datasets). For both the baselines and our model, we use ground truth camera intrinsics for model inference for all 3D datasets, except for SA1B-Hard which we used FOV estimated by MoGe-2~\cite{wang2025moge}.

We present the results in \Cref{tab:newdata}. Despite being trained on a large number of datasets, we find that prior work still struggle with these five domains, incurring a significant drop in performance. In contrast, our leave-one-out model shows strong generalization, owing to our more diverse data distribution and stronger training framework. Interestingly, we notice that existing methods constantly trade places for second across different datasets, reflecting strong dataset-specific biases. This indicates that each baseline overfit to a narrow slice of the underlying data distribution.

\begin{table*}[tb]
    \caption{Comparison on five new benchmark datasets. The best results are highlighted in bold, while the second-best results are underlined. MPJPE is computed on 24 SMPL keypoints.}
    \label{tab:newdata}
    \centering
    \resizebox{0.95\textwidth}{!}
    {
    \begin{NiceTabular}{clcccccccccccc}
    %\toprule
    & & \multicolumn{2}{c}{EE4D-Phy} & \multicolumn{2}{c}{EE4D-Proc} & \multicolumn{2}{c}{Harmony4D} & \multicolumn{2}{c}{Goliath} & \multicolumn{2}{c}{Synthetic} & \multicolumn{1}{c}{SA1B-Hard} \\
    \cmidrule(lr){3-4} \cmidrule(lr){5-6} \cmidrule(lr){7-8} \cmidrule(lr){9-10} \cmidrule(lr){11-12} \cmidrule(lr){13-13}
    & Models & \scriptsize{PVE $\downarrow$} & \scriptsize{MPJPE $\downarrow$}
             & \scriptsize{PVE $\downarrow$} & \scriptsize{MPJPE $\downarrow$}
             & \scriptsize{PVE $\downarrow$} & \scriptsize{MPJPE $\downarrow$}
             & \scriptsize{PVE $\downarrow$} & \scriptsize{MPJPE $\downarrow$}
             & \scriptsize{PVE $\downarrow$} & \scriptsize{MPJPE $\downarrow$}
             & \scriptsize{Avg-PCK $\uparrow$} \\
    \midrule
    & CameraHMR \cite{patel2024camerahmr} & \underline{71.1}  & \underline{58.8}  & \underline{70.3}  & \underline{60.2}  & \underline{84.6}  & \underline{70.8}  & 66.7  & \underline{54.5}  & 102.8 & 87.2  & 63.0   \\
    & PromptHMR \cite{wang2025prompthmr} & 74.6 & 63.4 & 72.0 & 62.6 & 91.9 & 78.0 & 67.2 & 56.5 & \underline{92.7} & \underline{80.7} & 59.0     \\
    & NLF \cite{sarandi2024nlf} & 75.9 & 68.5 & 85.4 & 77.7 & 97.3 & 84.9 & \underline{66.5} & 58.0 & 97.6 & 86.5 & \underline{66.5} \\
    \midrule
    & 3DB-H Leave-one-out (Ours) & \best{\textbf{49.7}} & \best{\textbf{44.3}} & \best{\textbf{52.9}} & \best{\textbf{47.4}} & \best{\textbf{63.5}} & \best{\textbf{54.0}} & \best{\textbf{54.2}} & \best{\textbf{46.5}} & \best{\textbf{85.6}} & \best{\textbf{75.5}} & \best{\textbf{73.1}} \\
    & 3DB-H Full dataset  (Ours) & \textcolor{gray}{{37.0}}  & \textcolor{gray}{{31.6}}  & \textcolor{gray}{{41.9}}  & \textcolor{gray}{{36.3}}  & \textcolor{gray}{{41.0}}  & \textcolor{gray}{{33.9}}  & \textcolor{gray}{{34.5}}  & \textcolor{gray}{{28.8}}  & \textcolor{gray}{{55.2}}  & \textcolor{gray}{{47.2}}  & \textcolor{gray}{{76.6}}  \\
    %\bottomrule
    \end{NiceTabular}
    }
\end{table*}

\begin{table}[tb]
\caption{Comparison on Freihand for hand pose estimation. Methods using Freihand for training are denoted by $^\dagger$.}
\label{tab:hand_result_freihand}
\centering
\resizebox{0.55\textwidth}{!}{
\begin{tabular}{lcccc}
Method & PA-MPVPE $\downarrow$ & PA-MPJPE $\downarrow$ & F@5 $\uparrow$& F@15 $\uparrow$\\
\midrule
LookMa~\cite{hewitt2024look} & 8.1 & 8.6 & 0.653 & - \\
METRO~\cite{Lin2020EndtoEndHP}$^\dagger$ & 6.3 & 6.5 & 0.731 & 0.984 \\ 
HaMeR~\cite{pavlakos2023hamer}$^\dagger$ & 5.7 & 6.0 & 0.785 & 0.990 \\
MaskHand~\cite{saleem2025maskhand}$^\dagger$ & 5.4 & 5.5 & 0.801 & 0.991\\ 
WiLoR~\cite{potamias2025wilor}$^\dagger$ & 5.1 & 5.5 & 0.825 & 0.993 \\
\midrule
3DB-H (Ours) & 6.3 & 5.5 & 0.735 & 0.988 \\
3DB-DINOv3 (Ours) & 6.2 & 5.5 & 0.737 & 0.988\\ 
\end{tabular}}
\end{table}

\subsection{Evaluating Hand Pose Estimation Performance}
One significant characteristic of \modelcode{} is its strong performance in estimating hand shape and pose. Previous full-body human pose estimation methods \cite{cai2023smplerx, baradel2024multihmr, lin2023onex} revealed a notable gap in hand pose accuracy compared to \textit{hand-only} pose estimation methods \cite{pavlakos2023hamer, potamias2025wilor}. This performance gap arises from two main factors. First, hand-only methods can leverage large-scale datasets of hand poses, whereas full-body methods cannot utilize these datasets because of the absence of full-body images and annotations. Second, a free-moving wrist allows hand pose models to more easily fit finger poses with 2D and 3D alignment, while for full-body methods, wrist rotation and position are highly constrained by the body's pose and position.
Despite these challenges, \modelcode{} demonstrates strong hand pose accuracy. \modelcode{} benefits from the flexible model training design that incorporates both hand and body data and the hand decoder. Additionally, being promptable, \modelcode{} provides a natural mechanism to align the wrists of the body prediction with those of the hands. 
%Limited by prohibitive data usage license, we cannot evaluate on most public full-body benchmarks. 
We evaluate \modelcode{}'s hand estimation on the representative FreiHand \cite{Freihand2019} benchmark in \Cref{tab:hand_result_freihand}. For fair comparison against hand-only models, we use the output from our hand decoder for evaluation. Despite not training on the Freihand dataset, which gives a strong in-domain boost, \modelcode{}'s hand pose estimation accuracy is already comparable to SoTA hand pose estimation methods that include Freihand alongside many other hand-centric datasets.

\begin{table}[tb]
    \caption{2D categorical performance analysis on the SA-1B Hard dataset.}
    \label{tab:2Dcategories}
    \vspace{-3mm}
    \centering
    \resizebox{0.90\textwidth}{!}{
    \begin{NiceTabular}{lcccc>{\columncolor{green!20}}c>{\columncolor{green!20}}c}
    %\toprule
    & \multicolumn{2}{c}{CameraHMR \cite{patel2024camerahmr}} & \multicolumn{2}{c}{PromptHMR \cite{wang2025prompthmr}} & \multicolumn{2}{c}{3DB} \\
    \cmidrule(lr){2-3} \cmidrule(lr){4-5} \cmidrule(lr){6-7}
    & \textsc{aPCK}(body) & \textsc{aPCK}(feet) & \textsc{aPCK}(body) & \textsc{aPCK}(feet) & \textsc{aPCK}(body) & \textsc{aPCK}(feet) \\
    \midrule
    Body\_shape - In-the-wild & 87.64 & 78.56 & 85.73 & 77.87 & \textbf{90.76} & \textbf{92.12} \\
    Camera\_view - Back or side view & 59.69 & 46.64 & 61.92 & 47.74 & \textbf{76.27} & \textbf{66.81} \\
    Camera\_view - Bottom-up view & 55.18 & 34.84 & 46.56 & 29.25 & \textbf{69.62} & \textbf{55.35} \\
    Camera\_view - Others & 51.48 & 33.80 & 54.39 & 38.55 & \textbf{76.62} & \textbf{71.52} \\
    Camera\_view - Overhead view & 55.08 & 39.46 & 43.65 & 24.63 & \textbf{73.33} & \textbf{66.94} \\
    
    Hand - Crossed or overlapped fingers & 73.20 & 62.85 & 72.48 & 62.43 & \textbf{81.36} & \textbf{84.04} \\
    Hand - Holding objects & 76.73 & 72.11 & 73.57 & 68.92 & \textbf{83.40} & \textbf{85.92} \\
    Hand - Self-occluded hands & 73.22 & 58.06 & 72.43 & 56.19 & \textbf{80.07} & \textbf{80.82} \\
    
    Multi\_people - Contact or interaction & 63.23 & 51.65 & 61.77 & 47.60 & \textbf{74.81} & \textbf{69.92} \\
    Multi\_people - Overlapped & 53.11 & 41.88 & 57.17 & 41.43 & \textbf{70.82} & \textbf{64.71} \\
    
    Pose - Contortion or bending & 47.08 & 32.78 & 42.61 & 20.98 & \textbf{65.20} & \textbf{53.04} \\
    Pose - Crossed legs & 63.95 & 32.24 & 56.15 & 27.35 & \textbf{76.40} & \textbf{58.80} \\
    Pose - Inverted body & 46.12 & 30.01 & 39.83 & 24.64 & \textbf{78.18} & \textbf{72.19} \\
    Pose - Leg or arm splits & 57.51 & 31.43 & 54.76 & 33.11 & \textbf{83.69} & \textbf{72.49} \\
    Pose - Lotus pose & 63.19 & 14.38 & 54.85 & 12.87 & \textbf{74.53} & \textbf{57.97} \\
    Pose - Lying down & 51.29 & 35.88 & 44.59 & 26.88 & \textbf{71.35} & \textbf{66.53} \\
    Pose - Sitting on or riding & 79.66 & 71.65 & 70.15 & 61.16 & \textbf{84.85} & \textbf{81.51} \\
    Pose - Sports or athletic activities & 78.93 & 69.34 & 73.62 & 60.37 & \textbf{85.10} & \textbf{82.80} \\
    Pose - Squatting or crouching or kneeling & 62.74 & 41.47 & 54.41 & 33.84 & \textbf{72.85} & \textbf{61.85} \\
    Visibility - Occlusion (foot cues) & 62.93 & 26.83 & 58.00 & 30.81 & \textbf{75.43} & \textbf{54.74} \\
    Visibility - Occlusion (hand cues) & 61.01 & 53.89 & 58.55 & 51.13 & \textbf{76.04} & \textbf{72.01} \\
    Visibility - Truncation (lower-body truncated) & 39.27 & - & 46.50 & - & \textbf{61.95} & - \\
    Visibility - Truncation (others) & 79.18 & 74.82 & 77.06 & 74.99 & \textbf{84.23} & \textbf{86.72} \\
    Visibility - Truncation (upper-body truncated) & 62.37 & 54.90 & 56.01 & 49.28 & \textbf{64.49} & \textbf{70.99} \\
    %\bottomrule
    \end{NiceTabular}
    }
\end{table}

\subsection{Evaluating 2D Categorical Performance}
To better understand the strengths and weaknesses of models on a variety of image types, we compare the performance across our 24 categories defined over SA1B-Hard~\cite{kirillov2023sa1b}. Our proposed evaluation set is designed to capture a broad spectrum of human appearance and activity in images, ensuring robust evaluation across real-world scenarios. It consists of 24 total categories, which are organized under several high-level groups: Body Shape, Camera View, Hand, Multi-person, Pose and Visibility.

\begin{figure}[!htbp]
    \centering
    \includegraphics[width=\linewidth]{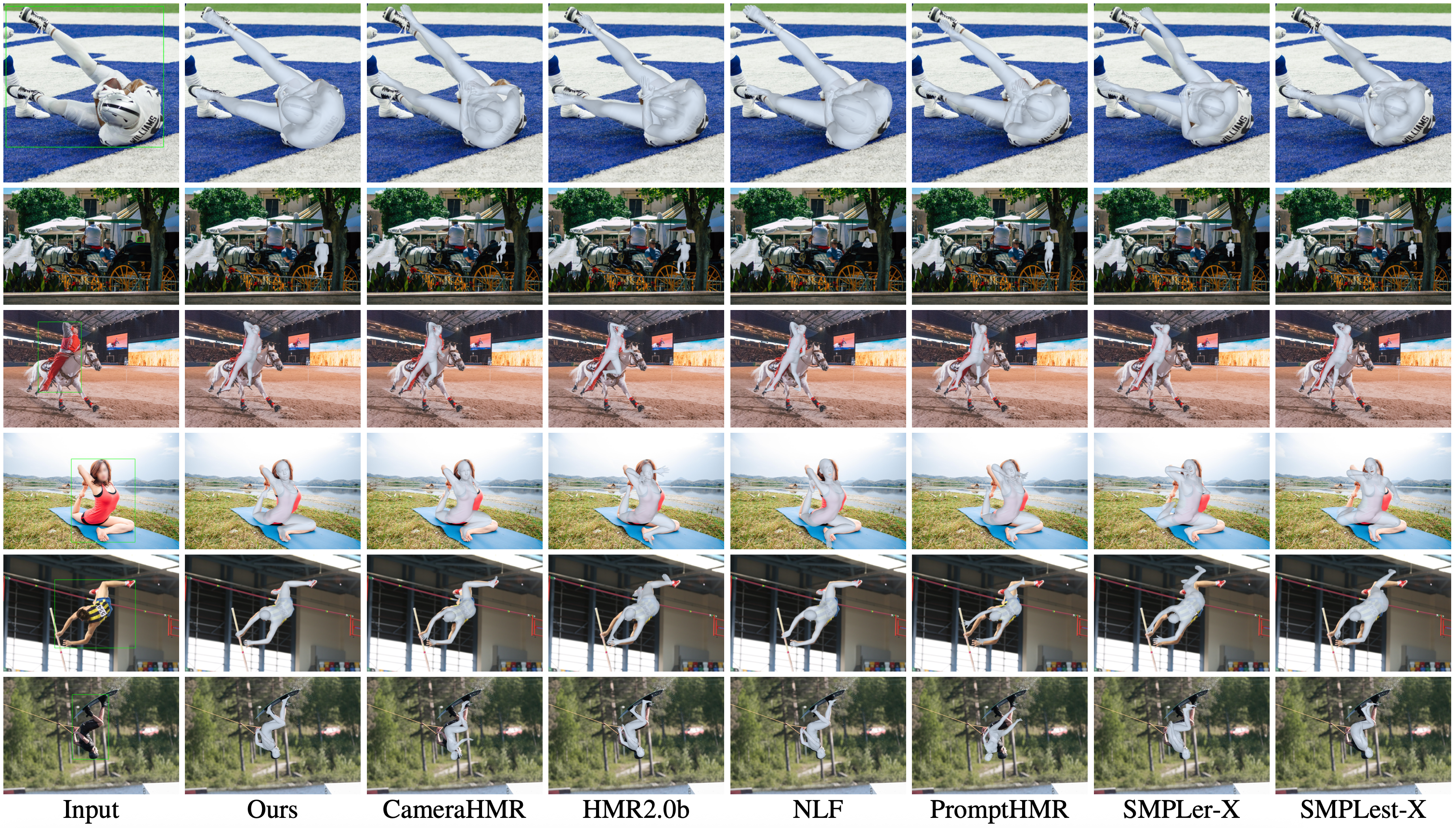}
    \caption{Qualitative comparison of \modelcode{} against state-of-the-art HMR methods. Source: SA-1B~\cite{kirillov2023sa1b}.}
    \label{fig:comp}
\end{figure}

We use the PCK (Percentage of Correct Keypoints) metric for 17 body keypoints and 6 feet keypoints. Results are reported using Avg-PCK, which is PCK averaged over a range of thresholds (\ie 0.01, 0.025, 0.05, 0.075, 0.1 of the human bounding box size). Results in \Cref{tab:2Dcategories} show that \modelcode{} outperforms all baselines on all categories. Qualitative examples are given in \Cref{fig:comp}.

One notable significance is for categories of \textit{Visibility - Truncation} where the model shows significant advantages than CameraHMR or PromptHMR.
Essentially, \modelcode{} has learned a much stronger pose prior when dealing with body truncation in images. Other rows with the large improvements are \textit{Pose - Inverted body} and \textit{Pose - Leg or arm splits}. We largely attribute these improvements to the increased distribution of hard poses selected by the data engine.

\begin{table}[tb]
    \caption{3D categorical performance analysis.}
    \label{tab:camera3db_nicetable_3d}
    \centering
    \resizebox{0.85\textwidth}{!}{
    \begin{NiceTabular}{lcccccc>{\columncolor{green!20}}c>{\columncolor{green!20}}c>{\columncolor{green!20}}c}
    %\toprule
    & \multicolumn{3}{c}{CameraHMR \cite{patel2024camerahmr}} & \multicolumn{3}{c}{PromptHMR \cite{wang2025prompthmr}} & \multicolumn{3}{c}{\modelcode{}} \\
    \cmidrule(lr){2-4} \cmidrule(lr){5-7} \cmidrule(lr){8-10}
    & PVE & MPJPE & PA-MPJPE & PVE & MPJPE & PA-MPJPE & PVE & MPJPE & PA-MPJPE \\
    \midrule
    aux:depth\_ambiguous           & 126.25 & 102.25 & 81.33 & 109.58 & 91.77 & 69.24 & \textbf{64.38} & \textbf{52.72} & \textbf{39.85} \\
    aux:orient\_ambiguous          & 84.26 & 71.77 & 45.07 & 83.79 & 72.93 & 46.17 & \textbf{42.35} & \textbf{36.64} & \textbf{25.16} \\
    
    aux:scale\_ambiguous           & 118.18 & 104.77 & 50.93 & 112.95 & 102.28 & 47.26 & \textbf{58.64} & \textbf{51.16} & \textbf{27.67} \\
    fov:medium                     & 82.88 & 68.81 & 46.86 & 76.31 & 64.84 & 42.85 & \textbf{43.58} & \textbf{36.97} & \textbf{25.57} \\
    fov:narrow                     & 82.15 & 69.82 & 49.73 & 90.41 & 77.95 & 53.49 & \textbf{52.14} & \textbf{43.89} & \textbf{36.18} \\
    
    fov:wide                       & 71.55 & 60.05 & 38.66 & 74.98 & 64.55 & 42.87 & \textbf{37.97} & \textbf{33.06} & \textbf{22.44} \\
    interaction:close\_interaction & 107.59 & 90.95 & 57.62 & 115.19 & 98.12 & 64.87 & \textbf{54.23} & \textbf{44.98} & \textbf{29.76} \\
    interaction:mild\_interaction  & 89.98 & 75.28 & 52.93 & 106.55 & 90.38 & 62.74 & \textbf{42.63} & \textbf{34.65} & \textbf{27.16} \\
    
    pose\_2d:hard                  & 117.91 & 107.74 & 77.16 & 117.73 & 110.64 & 79.16 & \textbf{62.93} & \textbf{57.50} & \textbf{45.58} \\
    pose\_2d:very\_hard            & 150.20 & 140.61 & 92.66 & 150.15 & 145.07 & 95.40 & \textbf{62.22} & \textbf{56.84} & \textbf{42.39} \\
    pose\_3d:hard                  & 133.89 & 121.11 & 84.21 & 129.30 & 118.59 & 81.82 & \textbf{71.42} & \textbf{63.68} & \textbf{49.10} \\
    
    pose\_3d:very\_hard            & 213.66 & 206.34 & 143.23 & 186.35 & 179.46 & 129.51 & \textbf{114.20} & \textbf{110.62} & \textbf{86.43} \\
    
    pose\_prior:average\_pose      & 68.52 & 56.70 & 37.22 & 70.32 & 59.73 & 39.42 & \textbf{36.06} & \textbf{30.95} & \textbf{21.35} \\
    pose\_prior:easy\_pose         & 57.83 & 47.31 & 29.92 & 62.85 & 53.58 & 32.80 & \textbf{29.53} & \textbf{24.66} & \textbf{17.20} \\
    pose\_prior:hard\_pose         & 94.64 & 80.04 & 54.53 & 88.12 & 76.19 & 51.15 & \textbf{51.65} & \textbf{44.24} & \textbf{31.09} \\
    shape:average\_bmi             & 70.35 & 58.07 & 38.08 & 71.01 & 60.25 & 39.90 & \textbf{36.58} & \textbf{31.41} & \textbf{21.31} \\
    
    shape:high\_bmi                & 84.52 & 69.96 & 47.55 & 79.49 & 67.83 & 43.04 & \textbf{43.33} & \textbf{36.49} & \textbf{22.45} \\
    shape:low\_bmi                 & 80.93 & 65.70 & 42.71 & 69.92 & 58.76 & 37.30 & \textbf{38.74} & \textbf{32.73} & \textbf{21.82} \\
    shape:very\_high\_bmi          & 87.18 & 72.91 & 47.54 & 81.17 & 69.05 & 44.03 & \textbf{48.51} & \textbf{41.11} & \textbf{24.80} \\
    shape:very\_low\_bmi           & 108.16 & 91.25 & 47.26 & 94.16 & 81.12 & 38.64 & \textbf{51.76} & \textbf{45.69} & \textbf{22.97} \\
    truncation:left\_body          & 135.30 & 113.17 & 87.98 & 127.53 & 110.67 & 91.33 & \textbf{91.28} & \textbf{76.46} & \textbf{62.23} \\
    
    truncation:lower\_body         & 127.81 & 97.84 & 75.82 & 151.52 & 118.65 & 83.79 & \textbf{92.87} & \textbf{67.10} & \textbf{60.77} \\
    truncation:right\_body         & 110.28 & 91.58 & 71.17 & 115.71 & 98.43 & 72.15 & \textbf{75.04} & \textbf{62.84} & \textbf{50.62} \\
    truncation:severe              & 230.51 & 213.64 & 124.01 & 186.57 & 168.22 & 122.70 & \textbf{126.53} & \textbf{113.66} & \textbf{88.42} \\
    truncation:upper\_body         & 85.59 & 79.68 & 56.36 & 86.06 & 80.88 & 56.94 & \textbf{50.83} & \textbf{48.79} & \textbf{38.39} \\
    viewpoint:average\_view        & 75.61 & 62.69 & 41.90 & 74.17 & 62.80 & 41.81 & \textbf{41.25} & \textbf{35.22} & \textbf{24.41} \\
    viewpoint:bottomup\_view       & 89.83 & 72.25 & 53.00 & 95.46 & 78.87 & 55.57 & \textbf{56.50} & \textbf{47.07} & \textbf{34.03} \\
    viewpoint:topdown\_view        & 101.69 & 91.13 & 59.15 & 104.29 & 97.92 & 63.39 & \textbf{42.84} & \textbf{38.78} & \textbf{27.90} \\
    %\bottomrule
    \end{NiceTabular}
    }
\end{table}

\subsection{Evaluating 3D Categorical Performance}

Categorical 3D analysis using existing single view datasets is challenging as the underlying pseudo ground truth are low-fidelity approximations of the real geometry. In order to perform a more detailed categorical analysis of HMR methods, we constructed an evaluation dataset using a mix of synthetic and real data from multi-view datasets with high camera counts (more than 100 cameras).

To comprehensively evaluate 3D human mesh reconstruction performance for HMR, we define a set of 28 distinct categories based on interpretable scene and subject attributes, such as occlusion, truncation, viewpoint, pose difficulty, shape, and interaction. Unlike the manual classification used for 2D categories, these 3D categories are automatically generated using rule-based criteria applied to metadata and geometric cues. This systematic approach enables consistent, scalable, and objective analysis of model performance across diverse real-world conditions.

Based on results from \Cref{tab:camera3db_nicetable_3d}, \modelcode{} demonstrates superior performance in challenging scenarios. Particularly within the \textit{very hard} pose categories, \modelcode{} consistently outperforms both CameraHMR and PromptHMR in the \textit{pose\_3d:very\_hard} category  and in \textit{pose\_2d:very\_hard}. These results indicate that \modelcode{} possesses inherent strengths in accurately estimating poses under the most challenging conditions.

Additionally, \modelcode{} exhibits a significant advantage in handling the \textit{truncation:severe} scenario in comparison to CameraHMR and achieves better performance in the \textit{viewpoint:topdown\_view} category in comparison to PromptHMR.

\begin{figure*}[t]
    \centering
    \includegraphics[width=0.9\linewidth]{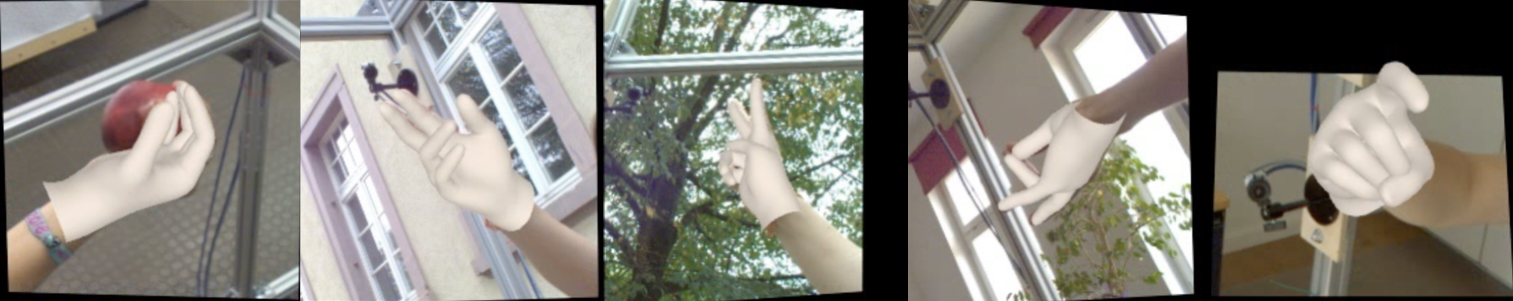}
    \caption{Qualitative results of hand estimation using the hand decoder of 3DB. Source: Freihand \cite{Freihand2019}.}
    \label{fig:freihand_qual}
\end{figure*}

\subsection{Qualitative Results}

In addition to quantitative gains, our model shows clear qualitative improvements over baselines. \Cref{fig:comp} compares \modelname{} to six state-of-the-art methods on the SA1B-Hard dataset, highlighting challenging cases with complex poses, shapes, and occlusions. As shown, \modelname{} consistently achieves more accurate body pose and shape recovery, especially for fine details like limbs and hands. The 2D overlays in \Cref{fig:comp} further illustrate better alignment with input images, demonstrating the robustness of our approach even under difficult conditions. When we focus on hand-crop images where the human body is invisible or truncated out of images, we demonstrate the effectiveness of model as in ~\Cref{fig:freihand_qual}. Here, we only visualize the mesh output by the hand decoder for simplicity and clearness. 

\subsection{Human Preference Study}

We conducted a large-scale user preference study to evaluate the perceptual quality of human reconstructions produced by \modelcode{} compared with existing approaches on the SA1B-Hard dataset. While quantitative metrics capture geometric and numeric accuracy, they do not always align with the human perception accuracy.

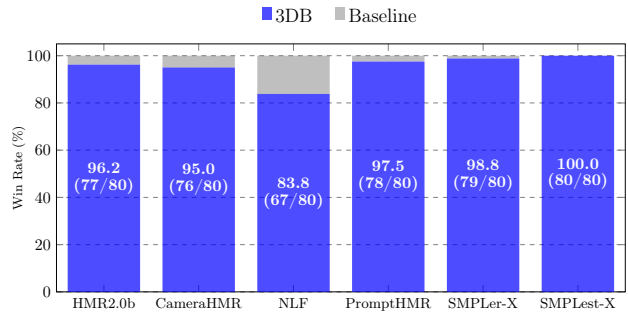
\begin{figure}[tb]
\centering
\resizebox{.5\columnwidth}{!}{%
   \begin{tikzpicture}
\begin{axis}[
    ybar stacked,
    bar width=45pt,
    width=14cm,
    height=7cm,
    ymin=0, ymax=105,
    ylabel={Win Rate (\%)},
    ylabel style= {font=\footnotesize, yshift=-6pt},
    symbolic x coords={HMR2.0b,CameraHMR,NLF,PromptHMR,SMPLer-X,SMPLest-X},
    xtick=data,
    xticklabel style={font=\small},
    enlarge x limits=0.1,
    legend style={
        at={(0.5,1.05)},
        anchor=south,
        legend columns=-1,
        draw=none,
        font=\large,
        /tikz/every even column/.append style={column sep=4pt},
    },
    legend image post style={draw opacity=0},
    ymajorgrids=true,
    grid style={dashed,draw=gray}
]

    \addplot+[
        fill={blue},
        fill opacity=0.7,
        draw=none,
        point meta=explicit symbolic,
        nodes near coords={\pgfplotspointmeta},
        nodes near coords align={center},
        every node near coord/.append style={
            font=\normalsize\bfseries,  % Increased size (was \tiny)
            text opacity=1,
            white
        },
    ] coordinates {
        (HMR2.0b,96.2)   [{\shortstack{96.2\\(77/80)}}]
        (CameraHMR,95.0) [{\shortstack{95.0\\(76/80)}}]
        (NLF,83.8)       [{\shortstack{83.8\\(67/80)}}]
        (PromptHMR,97.5) [{\shortstack{97.5\\(78/80)}}]
        (SMPLer-X,98.8)  [{\shortstack{98.8\\(79/80)}}]
        (SMPLest-X,100.0)[{\shortstack{100.0\\(80/80)}}]
    };

    \addplot+[
        fill={gray},
        fill opacity=0.5,
        draw=none,
    ] coordinates {
        (HMR2.0b,3.8)
        (CameraHMR,5.0)
        (NLF,16.2)
        (PromptHMR,2.5)
        (SMPLer-X,1.2)
        (SMPLest-X,0.0)
    };

    \legend{\modelcode{}~~~,Baseline} 
   \end{axis}
  \end{tikzpicture}
}
\caption{Comparison of \modelcode{} win rate against baselines for human preference study. Win rate (\%) and number of wins out of 80.}
\label{fig:sam3d-winrate}
\end{figure}

We designed six independent pairwise comparison studies, each comparing \modelcode{} against one baseline method: HMR2.0b~\cite{goel2023hmr20}, CameraHMR~\cite{patel2024camerahmr}, NLF~\cite{sarandi2024nlf}, PromptHMR~\cite{wang2025prompthmr}, SMPLer-X~\cite{cai2023smplerx}, and SMPLest-X~\cite{yin2025smplestx}. The study encompassed $7,800$ unique participants ($1,300$ unique per comparison) resulting in over $20,000$ total responses. Each participant was presented with a video stimuli. The left and right sides of the video displayed reconstructions from the two methods, and a video transition effect as used to fade-in the reconstruction result over the image. Participants were instructed to choose which 3D reconstruction better matched the original image by answering: \textit{``Which 3D model of the person better matches the original image, left or right?''}. We quantify results using win rate and vote share. Win rate is the percentage of stimuli for which \modelcode{} received more votes than the baseline. As summarized in ~\Cref{fig:sam3d-winrate}, \modelcode{} consistently outperforms all baselines. Focusing on the strongest baseline, NLF, \modelcode{} achieves a win rate of 83.8\%. 

\section{Conclusion}
\label{sec:conc}
We have presented \modelcode, a robust HMR model for body and hands. Our approach leverages the \meshname{} parametric body model, employs a flexible encoder–decoder architecture, and supports optional prompts such as 2D keypoints or masks to guide inference. A central advance of our work is in the supervision pipeline. Instead of relying on noisy monocular pseudo-ground-truth, we leverage multi-view capture systems, synthetic sources, and a scalable data engine that actively mines and annotates challenging samples. This strategy yields cleaner and more diverse training signals, supporting generalization beyond curated benchmarks. At the same time, \modelcode{} employs a separate hand decoder to enhance the hand pose estimation with hand crops as input which makes it comparable to SoTA hand pose estimation methods.

\section*{Acknowledgements}
We gratefully acknowledge the following individuals for their contributions and support: Vivian Lee, George Orlin, Nikhila Ravi, Andrew Westbury, Jyun-Ting Song, Zejia Weng, Xizi Zhang, Yuting Ye, Federica Bogo, Ronald Mallet, Ahmed Osman, Rawal Khirodkar, Javier Romero, Carsten Stoll, Shunsuke Saito, Jean-Charles Bazin, Sofien Bouaziz, Yuan Dong, Su Zhaoen, Alexander Richard, Michael Zollhoefer, Roman Radle, Sasha Mitts, Michelle Chan, Yael Yungster, Azita Shokrpour, Helen Klein, Mallika Malhotra, Ida Cheng, Eva Galper.

\clearpage
\newpage
\bibliographystyle{plainnat}
\bibliography{paper}

\clearpage
\newpage
\beginappendix

\section{Author Contributions}
SAM 3D Body represents a joint effort by the entire team -- all members contributed to paper writing and project release, with each author contributing to the following areas:

\noindent\textbf{Model:} Xitong Yang (model lead); Jinkun Cao (hand pose, model improvement); Jinhyung Park (MHR integration, model improvement); Nicolas Ugrinovic (multi-person interaction); Jiawei Liu (SAM 3D unification).

\noindent\textbf{Data:} Devansh Kukreja (data engine and infrastructure); Don Pinkus (manual annotation tooling); Taosha Fan (multi-view mesh fitting); Soyong Shin (single-view mesh fitting, dense keypoint detector); Jinhyun Park (MHR mesh fitting); Jinkun Cao (hand and whole-body data).

\noindent\textbf{Evaluation:} Xitong Yang (internal/external benchmarks); Jinkun Cao (hand pose evaluation); Jiawei Liu (human preference study, visualization); Nicolas Ugrinovic (mutli-person evaluation).

\noindent\textbf{Leadership and XFN:} Kris Kitani, Anushka Sagar, Piotr Dollar, Matt Feiszli, Jitendra Malik.

\section{Evaluating \modelcode{} Prompt Following}
\modelcode{} is a promptable model that supports conditioning on 2D keypoints or segmentation masks for controllable human pose estimation. In this section, we evaluate the model’s ability to follow the provided prompts and analyze their impact on pose estimation performance.

\noindent\textbf{2D Keypoint Prompt.} The 2D keypoint prompt provides a user-friendly mechanism for adjusting pose estimation by specifying joint locations in the image~\cite{liu2024ipose,yang2023neural}. In \Cref{tab:noise_ablation}, we present an ablation study on varying the number of keypoint prompts provided during inference, where the keypoint with the largest error is selected for prompting. We observe that the model effectively follows the prompts and both 2D and 3D performance improve as more prompts are provided (noise scale $=0$). Notably, although the keypoint prompt is provided in the 2D image space, \modelcode{} is able to leverage this information to infer a more accurate 3D body pose.

We further evaluate the sensitivity of our model to the quality of keypoint prompts, as shown in~\Cref{tab:noise_ablation} for $\#\text{Prompt}=1$. The noise scale is defined relative to the bounding box size, as in PCK. We observe that the model is robust to small keypoint inaccuracy (noise scale $<0.05$), as such noise naturally exists in annotations of in-the-wild datasets. When the noise level becomes larger, performance degrades because the model tends to follow the incorrect keypoint prompts.

\begin{figure}
    \centering
    \includegraphics[width=0.9\columnwidth]{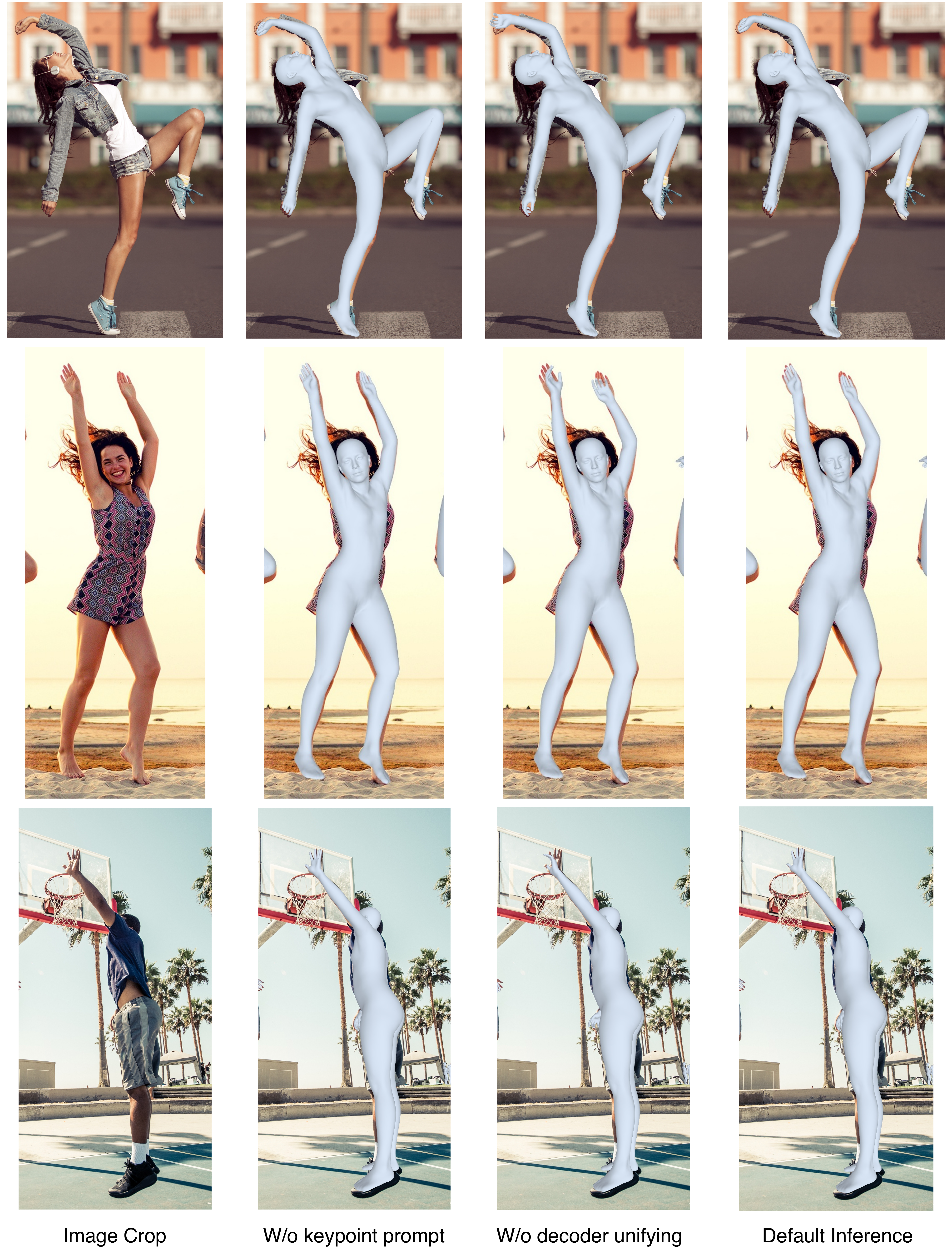}
    \caption{Qualitative comparison to show the impact from using keypoint prompting and unifying the predictions from hand decoder and body decoder.}
    \label{fig:ablation_qual}
\end{figure}

Finally, our full-body inference pipeline leverages the 2D keypoint prompting capability to improve hand pose estimation quality, as described in~\Cref{sec:inf}. To illustrate the impact of this strategy, we provide a qualitative comparison in~\Cref{fig:ablation_qual}. From this comparison, it is evident that without keypoint prompting, 2D keypoint alignment at the wrist and hand joints is significantly worse than that achieved by the default inference. On the other hand, without integrating the hand decoder during inference, the predicted wrist rotation is often suboptimal, leading to inferior 2D finger joint alignment.

\noindent\textbf{Maks Prompt.}
The capability of mask conditioning is essential when handling multiple people with close interaction, where the standard bounding box information is insufficient to clearly specify the person of interest for the model~\cite{yin2023hi4d,wang2025prompthmr}.
To assess the impact of incorporating masks as additional input to our model, we compare the model inference result with and without mask-conditioning on three multi-person (MP) datasets. We follow the prior work to provide ground-truth segmentation masks when available~\cite{yin2023hi4d,wang2025prompthmr}, and extract SAM2~\cite{ravi2025sam2} masks for the ITW dataset SA1B using the bounding boxes. As shown in \Cref{tab:mask_cond}, conditioning our model with person-specific segmentation masks yields significant improvements -- the same model (3DB-DINOv3) with mask conditioning improves PVE by 33.1 and MPJPE by 29.4 on Hi4D. Notably, both Hi4D~\cite{yin2023hi4d} and Harmony4D~\cite{khirodkar2024harmony4d} are multi-person dataset that captures close interactions between two individuals, featuring frames with significant occlusion, which poses a challenge for most HMR methods, especially in disambiguating between individuals. Using segmentation masks for each person as additional input, \modelcode{} effectively addresses this challenge and accurately predicts the corresponding person. For the experiments on our SA1B-Hard dataset, we observe that the performance gain on the "Multi-person" subset (+4.4\%) is more significant than that on the overall dataset (+0.9\%), indicating the importance of mask-conditioning for multi-person scenarios.

\begin{table}[t]
    \caption{Ablation on 2D keypoint prompting with 3DB-H. We report results under varying numbers of prompts, as well as different noise scales for a single prompt.}
    \label{tab:noise_ablation}
    \centering
    \scriptsize
    \resizebox{0.68\columnwidth}{!}{
    \begin{NiceTabular}{lccccccc}
    %\toprule
    \multicolumn{1}{c}{\# Prompts} 
        & \multicolumn{1}{c}{0} 
        & \multicolumn{5}{c}{1} 
        & \multicolumn{1}{c}{2} \\
    \cmidrule(lr){1-1} \cmidrule(lr){2-2} \cmidrule(lr){3-7} \cmidrule(lr){8-8}
    \multicolumn{1}{c}{Noise scale}
        & \multicolumn{1}{c}{0}
        & \multicolumn{1}{c}{0}
        & \multicolumn{1}{c}{0.01}
        & \multicolumn{1}{c}{0.03}
        & \multicolumn{1}{c}{0.05}
        & \multicolumn{1}{c}{0.1}
        & \multicolumn{1}{c}{0} \\
    \midrule
    COCO (\scriptsize{PCK@0.05$\uparrow$}) & 86.7 & \best{90.2} & 90.2 & 89.5 & 87.6 & 80.9 & \best{93.0} \\
    EMDB (\scriptsize{MPJPE $\downarrow$}) & 63.3 & \best{60.1} & 60.3 & 61.5 & 63.3 & 67.8 & \best{58.9} \\
    %\bottomrule
    \end{NiceTabular}
    }
\end{table}

\begin{table}[tb]
    \caption{Comparison on mask-conditioned inference with 3DB-DINOv3 on multi-person datasets.}
    \label{tab:mask_cond}
    \centering
    \scriptsize
    \resizebox{0.85\columnwidth}{!}{
    \begin{NiceTabular}{lcccccc}
    %\toprule
    & \multicolumn{2}{c}{Hi4D~\cite{yin2023hi4d}} & \multicolumn{2}{c}{Harmony4D} & \multicolumn{1}{c}{SA1B-Hard} & \multicolumn{1}{c}{SA1B-MP}  \\
    \cmidrule(lr){2-3} \cmidrule(lr){4-5} \cmidrule(lr){6-6} \cmidrule(lr){7-7} 
    \multicolumn{1}{c}{Models} & \scriptsize{PVE $\downarrow$} & \scriptsize{MPJPE $\downarrow$} 
             & \scriptsize{PVE $\downarrow$} & \scriptsize{MPJPE $\downarrow$} 
             & \scriptsize{Avg-PCK $\uparrow$}
             & \scriptsize{Avg-PCK $\uparrow$} \\
    \midrule
    3DB (w/o mask) & 91.4 & 76.4  & 42.7 & 35.6  & 75.4 & 67.9   \\
    3DB (w/ mask) & \best{58.3} & \best{47.0} & \best{36.5} & \best{30.1} & \best{76.3} & \best{72.3}    \\
    \end{NiceTabular}
    }
\end{table}

\section{Limitations}
We discuss some limitations of \modelcode{} as presented in this paper and suggest possible next steps to address these limitations.
First, \modelcode{} processes each individual separately, without taking multi-person or human-object interactions into account. This limits its ability to accurately interpret relative positions and physical interactions. A natural next step would be to incorporate interactions among humans, objects, and the environment into the model's training process.
Second, while our model has achieved significant improvements in hand pose estimation as part of the full-body estimation task, its accuracy does not surpass that of specialized hand-only pose estimation methods. Additionally, due to the limited availability of high-quality full-body data during training, the hand estimation performance from the body decoder alone is also suboptimal. This limitation can be addressed by incorporating more diverse full-body data into the training of \modelcode{}.
Third, both \modelcode{} and the underlying mesh model \meshcode{} fall short in modeling human body shapes across all age groups. As a result, they may produce suboptimal pose estimations and shape modeling for children.

\end{document}